# Heterogeneous Ensemble Learning for Enhanced Crash Forecasts – A Frequentist and Machine Learning based Stacking Framework

**Numan Ahmad**
Graduate Research Assistant
Department of Civil & Environmental Engineering
The University of Tennessee, TN 37996, USA
nahmad1@vols.utk.edu

**Dr. Behram Wali**
Lead Research Scientist,
Urban Design 4 Health, Inc.
bwali@ud4h.com

**Dr. Asad J. Khattak**
Beaman Distinguished Professor & Transportation Program Coordinator
Department of Civil & Environmental Engineering
The University of Tennessee, TN 37996, USA
akhattak@utk.edu

**ABSTRACT**

A variety of statistical and machine learning methods are used to model crash frequency on specific roadways – with machine learning methods generally having a higher prediction accuracy. Recently, heterogeneous ensemble methods (HEM), including “stacking,” have emerged as more accurate and robust intelligent techniques and are often used to solve pattern recognition problems by providing more reliable and accurate predictions. In this study, we apply one of the key HEM methods, “Stacking,” to model crash frequency on five-lane undivided segments (5T) of urban and suburban arterials. The prediction performance of “Stacking” is compared with parametric statistical models (Poisson and negative binomial) and three state-of-the-art machine learning techniques (Decision tree, random forest, and gradient boosting), each of which is termed as the base learner. By employing an optimal weight scheme to combine individual base learners through stacking, the problem of biased predictions in individual base-learners due to differences in specifications and prediction accuracies is avoided. Data including crash, traffic, and roadway inventory were collected and integrated from 2013 to 2017. The data are split into training, validation, and testing datasets. Estimation results of statistical models reveal that besides other factors; crashes increase with density (number per mile) of different types of driveways. Comparison of out-of-sample predictions of various models confirms the superiority of “Stacking” over the alternative methods considered. From a practical standpoint, “stacking” can enhance prediction accuracy (compared to using only one base learner with a particular specification). When applied systemically, stacking can help identify more appropriate countermeasures.

## 1. Introduction

Count data models (Poisson and negative binomial models) have been extensively used to model the relationships between crash frequency and key correlates, such as annual average daily traffic (AADT) and segment length, etc. (*1*-*3*). Compared to Poisson models, negative binomial variants are well-suited to capture potential over-dispersion in crash data. These models provide rich inferential insights into the mechanisms through which associated factors correlate with safety outcomes. However, given the intrinsic parametric nature of the models and the subsequent assumptions, prediction accuracy of count data models is often a concern. Growing evidence of the role of more accurate crash predictions in designing more appropriate safety countermeasures has led to an increased interest in machine learning (ML) methods. Unlike count data models, ML methods do not place strong restrictions on the specifications of the model (*4*). ML methods are more adequate for modeling complex non-linear relationships that frequently arise in crash data modeling. Tree-based regression (TBR) is one of the most popular and widely-used ML methods that does not require variable transformations and parametric assumptions (*5; 6*). TBR determines significant non-linear relationships among various predictor variables as well as computes the relative influence of predictors on response outcome (*5; 6*). However, the TBR technique is prone to instability leading to estimation results with higher variance (*5*). Ensemble methods like random forest regression (RFR) and gradient boosting regression (GBR) combine the estimates of numerous trees compared to a single tree, leading to improved stability and prediction accuracy (*5*). GBR technique ensembles numerous trees in a sequential way with a slower learning rate that captures a higher variance in data compared to the RFR method (*5*). While the prediction accuracy of ML methods usually is greater than the count data models, it lacks a holistic inferential framework providing little to no information about the safety mechanisms that link unsafe outcomes with key risk factors. Also, almost all of the ML methods explicitly relate to the bias-variance trade-off contour with different methods minimizing bias or variance. There is no escaping the relationship between bias and variance in ML models. Thus, the use of the single supervised or the unsupervised ML method could lead to relatively less accurate predictions.

While traditional count data models and ML methods have been extensively used in the safety literature, studies that combine the predictive (and inferential) strengths of both paradigms or the strengths of multiple ML methods are rare. The prediction performance of ML methods can be further improved by using more robust and heterogeneous ensemble methods (HEM), such as composite systems, stacking, or blending (*7; 8*). HEMs including "stacking" have emerged as more accurate and reliable intelligent techniques in pattern recognition issues. The idea of "Stacking" essentially helps in harnessing the gains simultaneously from less biased and low-variance predictions offered by different ML methods. For example, the GBR method builds on so-called "weak classifiers" - reducing prediction error mainly by reducing bias (and to some extent variance, by aggregating the predictions from many trees). Through heterogeneous ensemble methods such as "Stacking", predictive gains from different methodologies can be combined. Note that ensembles including RFR, GBR, and stacking are used to improve out-of-sample prediction accuracy and can be classified into (i) Homogeneous ensemble, and (ii) Heterogeneous ensemble (*9*). The homogeneous ensemble (e.g., RFR and GBR) uses the same feature selection algorithm with different training or learning datasets distributed over various nodes (*10*). Instead, the heterogeneous ensemble (i.e., stacking) uses different feature selection algorithms (e.g., Poisson, Negative binomial, TBR, RFR, and GBR) where the stacking meta-learner (which can be any statistical or ML method) blends the optimal combinations of predictions by base-learners and acts as a single decision-maker in the second-stage (*9*). Both homogeneous and heterogeneous ensembles can be used in regression as well as classification contexts. Compared to homogeneous ensembles, heterogeneous ensembles typically show significant performance gains. Studies suggest that heterogeneous ensembles do not only outperform the conventional statistical models and other ML methods but also show superior prediction performance compared to homogeneous ensembles (*11*). In transportation safety, homogeneous ensembles have been widely used for predicting crash frequency (*12*) and severity given a crash (*13*). Some studies used heterogeneous ensembles (e.g., stacking) in classification context to predict injury severity (*14*). However, the application of heterogeneous ensemble (stacking) to predict crash frequency on roadways has not been or very lightly explored to the best of the authors' knowledge. Given the prevalent gaps in the literature discussed above, this study contributes by:

- Applying a rigorous and robust HEM scheme to model and predict crash frequency on five-lane (5T) undivided segments on urban and suburban arterials, including two-way left-turn lanes (2WLTL)
- Comparing the prediction performance of "Stacking" with traditional statistical models and three state-of-the-art machine learning techniques (decision trees, random forest, and gradient boosting regression).

## 2. Methodology

### 2.1. Conceptual Architecture: Heterogeneous Ensemble Methods (Stacking)

Stacking generally provides higher prediction accuracy compared to base-learners (*15*). Suppose $Y$ is the response outcome, $X$ is the set of predictors used in individual models (briefly discussed in subsequent sections), and $g_1, g_2, \ldots, g_L$ are the predictions obtained using base learners (*15*). The prediction function for the linear ensemble (stacked) model can be given as (*15*):

$$b(g) = (w_1 * g_1) + (w_2 * g_2) + \cdots + (w_L * g_L) \quad (1)$$

Note that $w_i$ indicates the weight assigned to an individual model in the stacking technique (*15*). The model weights ($w_i$) are used to minimize MSE between actual response variable ($y_i$) and prediction outcome of meta-learner (stacked ensemble technique) as shown (*15*):

$$\min \sum_{i=1}^{N} (y_i - (w_1 * g_{1i} + w_2 * g_{2i} + \cdots + w_1 * g_{Li}))^2 \quad (2)$$

The conceptual design of this study is presented in Figure 1. First, a randomly selected subsample containing 304 roadway segments of 5T urban and suburban arterials for a period of five years (2013-2017) was selected for analysis. Next, we split data into training (2013-2015), validation (2016), and testing (2017) datasets (Figure 1). Note that in all the three datasets, only crashes and average annual daily traffic may change while all other factors remain the same. We follow this splitting procedure to develop a crash prediction model which can be reused with updated data to forecast crashes in the future. First, five individual base-learners are trained using training data to model crash frequency per year (2013-2015). Next, prediction outcomes obtained from these five base-learners using the validation dataset are obtained and combined with actual crashes reported in 2016, which generates a new training dataset for the meta-learner (stacking). The results are finally averaged to get a single estimation. Finally, we apply individual base learners (trained using the training dataset) and meta-learners or the stacked model (trained using validation dataset) to the new data (2017) to accurately compare their prediction performance (Figure 1).

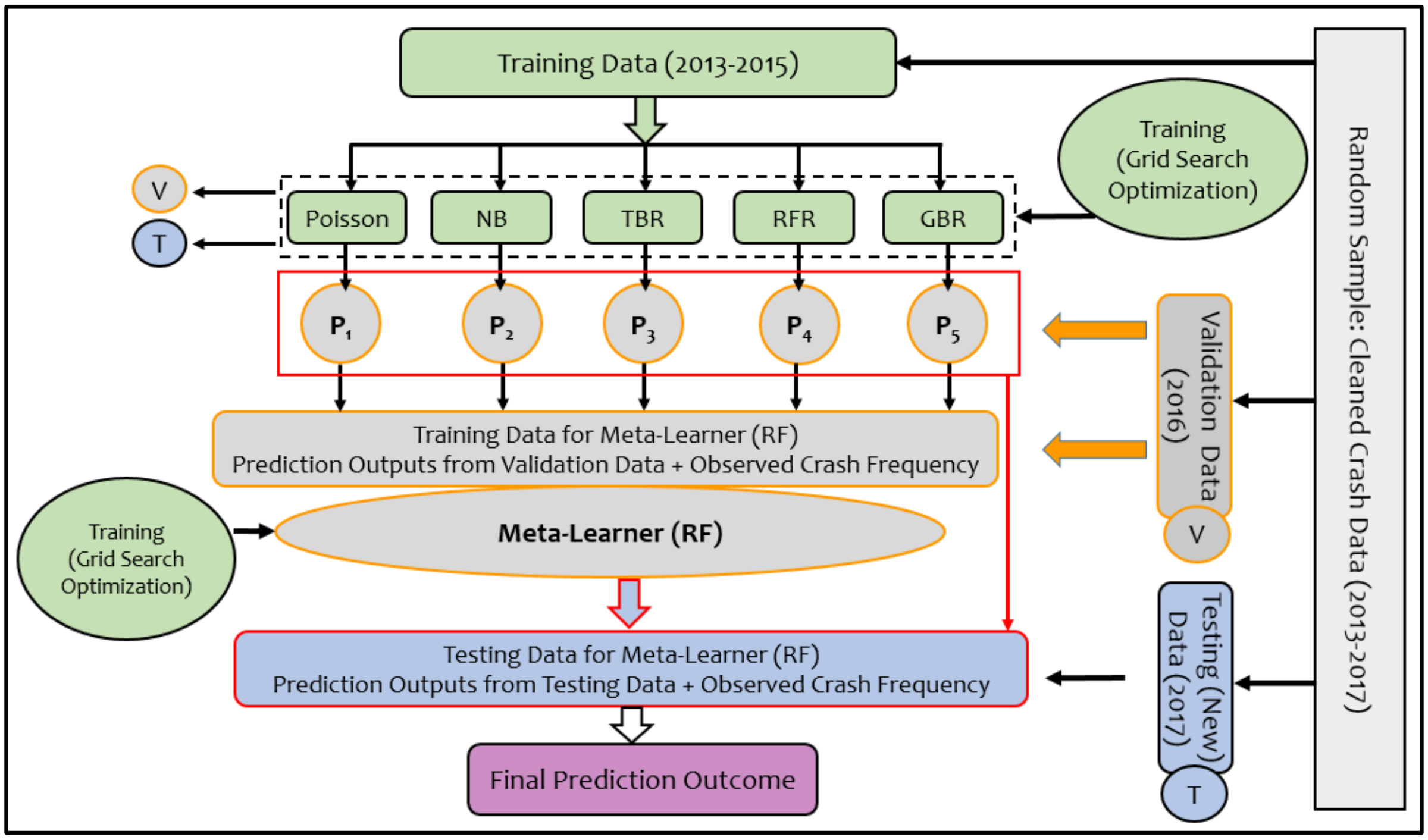

**Figure 1. Conceptual Design of Stacking Ensemble Utilized for Crash Frequency Modeling**

**Notes:** In Figure 1, NB indicates a negative binomial model, TBR indicates a Tree-based regression, RFR indicates an RFR, and GBR indicates a gradient boosting regression. $P_1$, $P_2$, $P_3$, $P_4$, and $P_5$ are the prediction outcomes obtained while applying Poisson, NB, TBR, RFR, and GBR models to the validation dataset, respectively. V and T indicate validation and testing datasets, respectively.

This study applies stacking where a meta learner is used to combine multiple predictions obtained from various base learners, as explained below.

- **Base Learner:** Any statistical model or ML method when applied in the first stage of stacking is termed as a "base learner" in this study. For instance, this study applies five base learners which include two statistical models (Poisson and Negative binomial) and three ML methods (TBR, RFR, and GBR).
- **Meta Learner:** The stacking meta-learner algorithm is an ensemble technique that combines predictions from two or more than two base-learners specifically to further enhance prediction accuracy. This study uses three ML methods including TBR, RFR, and GBR as meta-learners to combine predictions for the five base-learners (Poisson, negative binomial, TBR, RFR, and GBR). Note that stacking is termed as a "heterogeneous ensemble" that combines different feature selection procedures (Poisson, Negative binomial, TBR, RFR, and GBR).

## 2.2. Count Data Models: Poisson and Negative Binomial Regression

Studies suggest using count data (Poisson and negative binomial regression) models to explore the relationship of the crash frequency with explanatory variables (*17*). The mathematical formula of Poisson regression can be given below (*18*).

$$\mathrm{P}(n_i) = \frac{exp\ (-\lambda_i)\lambda_i^n}{n_i!} \quad (3)$$

Where $P(n_i)$ is the probability of a crash occurring on a specific road segment $(i)$, $(n)$ is the frequency of a crash on a specific road segment at a particular time, and $(\lambda_i)$ is the expected number of crashes occurring on a particular road segment $(i)$ in a specific duration. The expected number of crashes $(\lambda_i)$ is linked to its key contributing factors as below (*18*):

$$ln(\lambda_i) = \beta(X_i) \quad (4)$$

Where $X_i$ indicates a set of explanatory variables, and $\beta$ are their associated parameter estimates.

In the case of over-dispersion, Poisson regression is not preferable due to violation of its basic assumption, therefore negative binomial regression is suggested as below (*18*).

$$ln(\lambda_i) = \beta(X_i) + \in_i \quad (5)$$

Where $exp\ (\in_i)$ is an error term with gamma distribution "mean equals one and variance $(\alpha)$" (*18*).

## 2.3. Machine Learning Methods

### *2.3.1. Decision-tree Regression*

The TBR uses a fast algorithm that recursively splits training data into smaller subsets (*19*). However, instability and reliability issues are key weaknesses of TBR (*19*). The algorithm searches to determine a splitting point with the lowest value of mean square error (MSE). At the optimal splitting point, the parent node is further split into two child nodes and the process continues until the optimal tree length is determined (reducing impurity associated with terminal node). The algorithm chooses the best splitter $(S^*)$ considering deviance $(D)$ or MSE at a node as:

$$D(t) = \sum_{x \in t}(Y_n - \hat{\mu})^2 \quad (6)$$

Where, $\hat{\mu}$ is a sample mean $(\bar{y})$ or mean estimate, $t$ indicates a specific node, and $X$ indicates a set of predictors. Referring to the generalized linear models, deviance $(D)$ is also termed as log-likelihood ratio statistics and can be written as:

$$D = 2 * l(\mu_{max}; y) - l(\hat{\mu}: y) \quad (7)$$

Where, $\mu_{max}$ is the maximum likelihood estimate.

### *2.3.2. Random Forest Regression*

Studies suggest ensemble methods like RFR and GBR to mitigate instability issues related to a single decision tree (*20*). The RFR algorithm works on a similar principle to the single decision tree; however, the key difference is that RFR assembles an enormous number of trees. The RFR algorithm selects a predictor at each node to maximize homogeneity at successive nodes (*20*). Regularization parameters considered for RFR include (*20*):

- Number of predictors selected at each node for split-up $(\boldsymbol{m_{try}})$
- Number of trees in forest $(n_{tree})$
- Number of maximum nodes in the forest

To determine the optimal value of $\boldsymbol{m_{try}}$, we use an extended grid-search optimization and 10-fold cross-validation procedure. To select an optimal pair of $n_{tree}$ and $\boldsymbol{m_{try}}$, two performance criteria including MSE and $R^2$ values are usually used:

$$MSE \approx MSE_{OOB} = \frac{1}{n}\sum_{i\in OOB}(y_i - \hat{y}_i)^2 \qquad (8)$$

$$R^2 = 1 - \frac{MSE_{OOB}}{Var(y_i)} \qquad (9)$$

Where $MSE_{OOB}$ is the MSE for the out-of-bag $(OOB)$ sample, $y_i$ is the observed number of crashes occurring on $i^{th}$ roadway segment in $OOB$ sample, $\hat{y}_i$ is predicted crashes on $i^{th}$ road segment in $OOB$ sample, $n$ is number of roadway segments in $OOB$ sample, $Var(y_i)$ is the variance of response outcomes $(y)$ determined as $\frac{1}{n}\sum_{i\in OOB}\ (y_i - \overline{y})^2$, while $\overline{y}$ is mean value of $y_i$ in the $OOB$ sample.

### *2.3.3. Gradient Boosting Regression*

Similar to the RFR approach, GBR is a pool procedure to enhance prediction accuracy. The algorithm calculates residuals after fitting the first tree due to which the GBR algorithm assigns more weight to such observations while fitting the next tree and so on (*5*). Let $f(x)$ be an approximation function of response outcome $(y)$ as predicted by a set of predictor variables $(x)$. In GBR approach, an additive expansion of the basic functions $(x\colon\gamma_m)$ can be given as:

$$f(x) = \sum_m f_m(x) = \sum_m \beta_m b(x\colon\gamma_m) \qquad (10)$$

Note that $\beta_m (m = 1,2,3, \dots, M)$ indicates the expansion coefficients, $b(x\colon\gamma_m)$ indicates single regression trees having parameter $(\gamma_m)$ as a split variable, and $\beta_m$ are the weights assigned to every tree (*5*). The algorithm estimates parameters like $\beta_m$ and $\gamma_m$ to minimize loss function $L(y(f(s))$ indicating prediction performance in term of deviance (*5*). Note that while GBR may nicely fit to the data, it can also lead to overfitting (*5*). To cure this issue, studies suggest selecting appropriate regularization parameters including the number of trees, shrinkage (learning rate), and complexity which helps in achieving a balance between variance and bias (*5*).

## 2.4. Model Performance

To evaluate the prediction performance of individual models (base learners) and stacked regression, we compare their Root Mean Square Error (RMSE) and Mean Absolute Error (MAE) based on the testing dataset (2017):

$$RMSE = \sqrt{\frac{1}{n}\sum_{i=1}^{n}(f_i - y_i)^2} \qquad (11)$$

$$MAE = \frac{1}{n}\sum_{i=1}^{n}|f_i - y_i| = \frac{1}{n}\sum_{i=1}^{n}|e| \qquad (12)$$

The value of $n$ is the total number of roadway segments, and $f_i$ and $y_i$ indicate predicted and observed crash frequency, respectively. Low values of RMSE and MAE indicate higher prediction accuracy.

# 3. Results and Discussion

## 3.1. Data Processing and Descriptive Statistics

The Enhanced Tennessee Roadway Information Management System (ETRIMS), maintained by the Tennessee Department of Transportation (TDOT), showed a total of 3,208 (753.97 miles) segments of state-maintained 5T urban and suburban arterials in Tennessee (shown in Figure 2) which was first cleaned and then a random sample (N = 304) was selected for analysis.

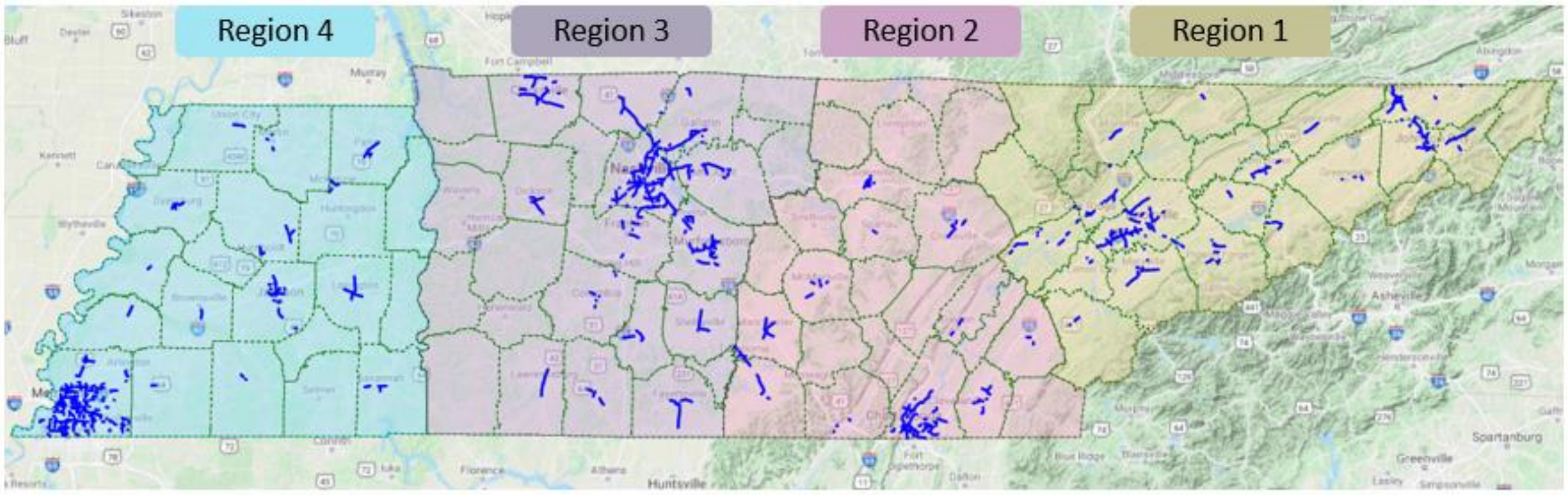

**Figure 2.** Distribution of the Overall 5T roadway segments of Urban and Suburban arterials in TN

**Note:** Tennessee has 95 counties, which are divided into four TDOT regions, shown on the map.

Following the HSM guidelines (*21*), segments shorter than 0.1 miles were removed leading to a reduced dataset containing 1,519 segments (totaling 523.93 miles). First, we determined the sample size to be selected from the population (1,519 segments) using 95% confidence level criteria. A random sample of 317 segments (105.78 miles) was selected for which crash (2013-2017), roadway geometry, and traffic data (2013-2017) were extracted using ETRIMS and TDOT Traffic History Application. Finally, 304 (103.27 miles) segments with complete data are analyzed. The distribution of roadway segments of 5T urban and suburban arterials (random sample "N = 304") in Tennessee based on the total number of crashes that have occurred on these segments from 2013 to 2017 is shown (Figure 3). The segments with a higher number of total crashes during the five years are mostly located in Region 3, which contains the Nashville area. Notably, segments with a low number of crashes over the 5 years are mostly located in the suburbs of the major cities and small cities.

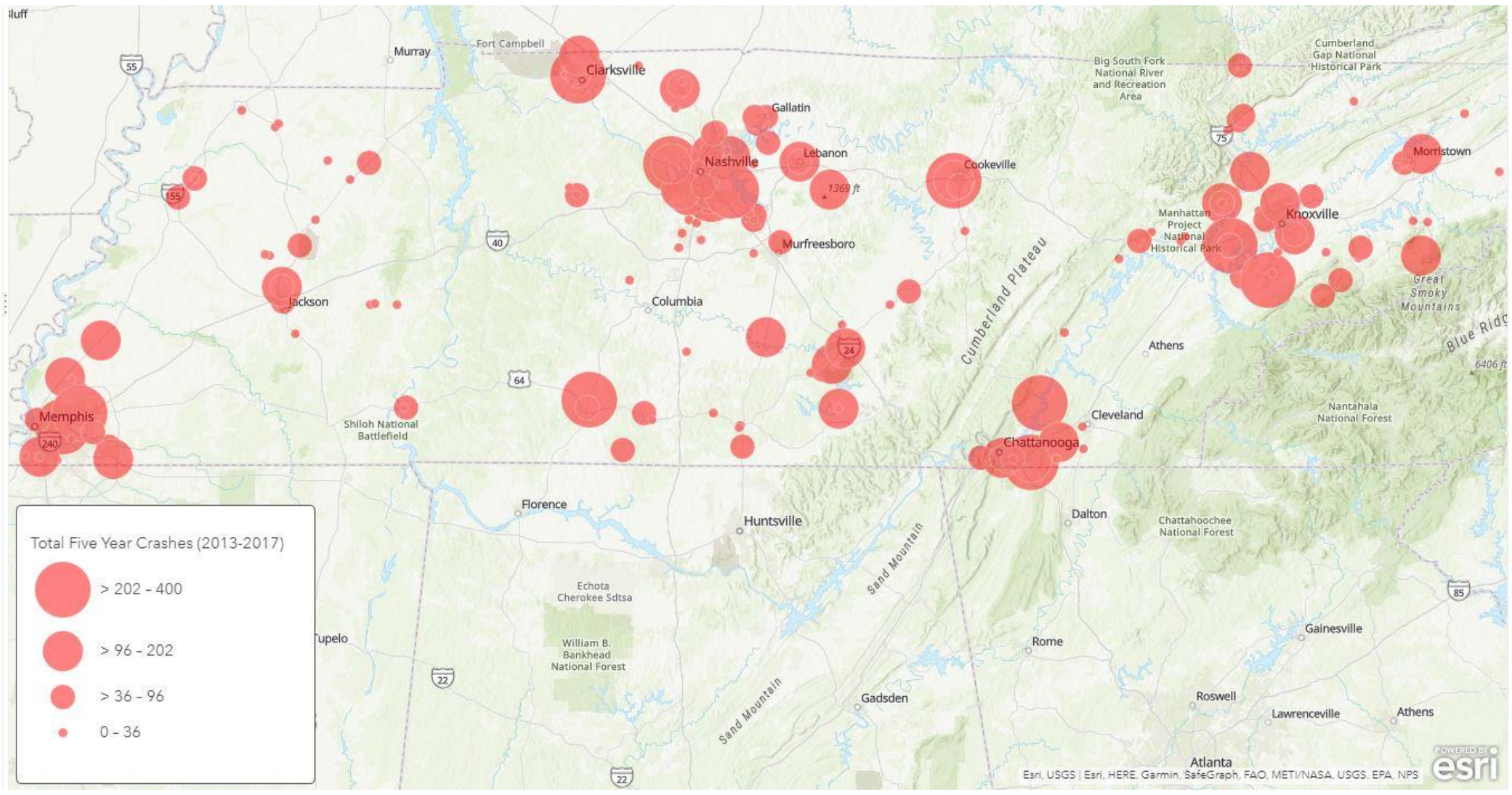

**Figure 3.** Distribution of 5T Urban and Suburban Arterial Segments based on Five-Year (2013-2017) Crashes in TN.

To achieve our study objective, the data is split into three subsets: training, validation, and testing. Table 1 presents descriptive statistics of key variables. Statistics reveal an average of 11.026 crashes (standard deviation of 14.020) across the three years on 5T segments of urban and suburban arterials. Crash distributions for validation (2016) and testing (2017) samples in Table 1 reveal similar distributions across the three (training, validation, testing) streams. Statistics for traffic measures and roadway geometric features are provided in Table 1. In 2017, the mean AADT (in 1000s) was 19.903 which is slightly higher than the yearly AADT in 2016 and average AADT per year from 2013 to 2015. This shows that on average, AADT per year has increased slightly compared to the previous years. The sample statistics for segment length and other key variables are shown in Table 1. The descriptive statistics seem reasonable because the dataset contains little to no outliers.

**Table 1. Descriptive Statistics of Key Variables: 5T Segments of Urban and Suburban Arterials**

| Variables | Obs. | Mean | Std. Dev. | Min | Max |
|---|---|---|---|---|---|
| Average Three-years Crashes (2013-15) | 304 | 11.026 | 14.020 | 0.000 | 72.000 |
| Total Crashes (2017) | 304 | 11.072 | 14.618 | 0.000 | 100.000 |
| Total Crashes (2016) | 304 | 11.010 | 14.651 | 0.000 | 90.000 |
| Average Annual Daily Traffic (AADT) per Year (2013-15) in 1000s | 304 | 19.098 | 8.938 | 3.182 | 49.766 |
| Average Annual Daily Traffic (AADT) (2017) in 1000s | 304 | 19.903 | 9.154 | 3.811 | 54.564 |
| Average Annual Daily Traffic (AADT) (2016) in 1000s | 304 | 19.644 | 9.210 | 3.613 | 54.360 |
| Segment length (mile) | 304 | 0.340 | 0.279 | 0.100 | 1.809 |
| Density (frequency per mile) of Major Commercial Driveways | 304 | 0.349 | 0.781 | 0.000 | 6.000 |
| Density (frequency per mile) of Minor Commercial Driveways | 304 | 0.865 | 1.626 | 0.000 | 12.000 |
| Density (frequency per mile) of Major Industrial Driveways | 304 | 0.461 | 0.936 | 0.000 | 7.000 |
| Density (frequency per mile) of Minor Industrial Driveways | 304 | 1.286 | 1.844 | 0.000 | 11.000 |
| Average Offset Distance (feet) to Roadside fixed objects | 304 | 14.266 | 8.188 | 0.000 | 30.000 |

### 3.2. Estimation Results

#### 3.2.1. Count Data Models: Poisson and Negative Binomial Regression

As a first step, we apply Poisson and negative binomial models to explore the average three-years (2013-2015) crash frequency. Both models come from a series of trials evaluated based on statistical significance, parsimony, and intuition. To select more appropriate models (with superior fit), several trials were made based on the specifications of explanatory variables. Poisson and Negative Binomial models with log-transformed AADT and segment length variables (Model 3 and Model 4) outperformed the counterparts with untransformed variables based on AIC, BIC, and log-likelihood values at convergence. To understand the relationship of key variables and crash frequency, we discuss the marginal effects (MEs) of variables for Model 4 which has the best in-sample fit. Our findings indicate that yearly crash frequency increases by almost 13 units with a unit increase in yearly AADT in 1000s (ln form) while keeping all other variables at their means (Table 2). Similarly, a unit increase in segment length (ln form) is associated with increases in crash frequency by 5.81, while keeping other variables at their mean values (Table 2). The estimation results of the best-fit model suggest that major commercial driveways have a stronger association compared to other types of driveways with crash frequency i.e., a unit increase in density of major commercial driveways associates with an increase in yearly crashes by 1.135 units. Moreover, yearly crash frequency is higher by 1.078, 0.744, and 0.505 with a unit increase in density of minor commercial driveways, major industrial driveways, and minor industrial driveways. Other studies suggest similar findings (*21*). These findings were expected as an increase in commercial and industrial driveways increases potential conflict points and creates a potential for gap acceptance errors. These findings highlight the need for investigating proactive access management strategies which can potentially reduce crashes specifically on 5T roadway segments of urban and suburban arterials. Our findings indicate that higher offset distance to roadside fixed objects is associated with fewer crashes. This was expected as fixed objects (i.e., utility pole, traffic sign, tree, and billboards) along roadway segments are potential safety risks specifically for errant vehicles (*22*). Such objects are more prevalent along with urban roadway segments (*22*).

**Table 2. Estimation Results of Poisson and Negative Binomial Models**

| **Variables** | **Poisson (Model 1)** | | | **Negative Binomial (Model 2)** | | | **Poisson (Model 3)** | | | **Negative Binomial (Model 4)** | | |
|---|---|---|---|---|---|---|---|---|---|---|---|---|
| | Data (2013-2015) | | | Data (2013-2015) | | | Data (2013-2015) | | | Data (2013-2015) | | |
| | Coeff. | t-stat | MEs | Coeff. | t-stat | MEs | Coeff. | t-stat | MEs | Coeff. | t-stat | MEs |
| Constant | 0.8804 | 13.39 | --- | 0.5274 | 2.96 | --- | -0.527 | -3.44 | --- | -0.446 | -1.28 | --- |
| Average Annual Daily Traffic (AADT) per Year (2013-15) in 1000s | 0.0548 | 30.49 | 0.603 | 0.0603 | 10.41 | 0.707 | --- | --- | --- | --- | --- | --- |
| Segment length (mile) | 0.8223 | 13.85 | 9.066 | 0.9078 | 4.17 | 10.656 | --- | --- | --- | --- | --- | --- |
| Density (frequency per mile) of Major Commercial Driveways | 0.1156 | 6.87 | 1.274 | 0.1298 | 1.94 | 1.524 | 0.072 | 4.32 | 0.795 | 0.102 | 1.54 | 1.135 |
| Density (frequency per mile) of Minor Commercial Driveways | 0.0808 | 8.80 | 0.891 | 0.1183 | 3.53 | 1.388 | 0.063 | 7.03 | 0.693 | 0.097 | 2.91 | 1.078 |
| Density (frequency per mile) of Major Industrial Driveways | 0.0765 | 5.28 | 0.843 | 0.1163 | 2.07 | 1.364 | 0.045 | 3.09 | 0.496 | 0.067 | 1.22 | 0.744 |
| Density (frequency per mile) of Minor Industrial Driveways | 0.0495 | 5.65 | 0.545 | 0.0675 | 2.21 | 0.792 | 0.029 | 3.41 | 0.321 | 0.045 | 1.50 | 0.505 |
| Average Offset Distance (feet) to Roadside fixed objects | -0.0215 | -8.05 | -0.237 | -0.0148 | -2.18 | -0.173 | -0.028 | -10.12 | -0.303 | -0.019 | -2.84 | -0.211 |
| **Key Variables (ln form)** | | | | | | | | | | | | |
| AADT per Year (2013-15) in 1000s (ln form) | --- | --- | --- | --- | --- | --- | 1.261 | 28.72 | 13.904 | 1.146 | 11.42 | 12.756 |
| Segment length (mile) (ln form) | --- | --- | --- | --- | --- | --- | 0.567 | 18.03 | 6.250 | 0.522 | 5.87 | 5.805 |
| **Over-dispersion Parameter** | --- | --- | --- | 0.5581 | 9.63 | --- | --- | --- | --- | 0.528 | 9.49 | --- |
| **Summary** | | | | | | | | | | | | |
| Sample Size | 304 | | | 304 | | | 304 | | | 304 | | |
| Log likelihood at Convergence | -1470.852 | | | -938.271 | | | -1395.448 | | | -931.1744 | | |
| AIC | 2957.704 | | | 1894.541 | | | 2806.896 | | | 1880.349 | | |
| BIC | 2987.44 | | | 1927.994 | | | 2836.632 | | | 1913.802 | | |

**Note:** AIC is Akaike Information Criterion, BIC is Bayesian Information Criterion, while MEs indicate marginal effects.

### 3.2.2. Machine Learning Techniques

#### *3.2.2.1. Single Decision Tree Regression*

First, we apply single TBR to predict average crash frequency per year on 5T urban and suburban arterials using a training dataset (2013-2015). Using one standard-error rule, we do not observe a significant reduction in error after a tree size of 7 (with cost complexity ~ 0.01858831). Using the mentioned optimal values of tuning parameters, an optimal tree is grown as shown on the right side in Figure 4. The key predictor variables used in developing the optimal tree include AADT (2013-2015) per year (ln form), segment length "mile" (ln form), the density of minor commercial driveways, and density of major commercial driveways (Figure 4). Note that the single decision tree is easily interpretable. For instance, it can be seen that if AADT (2013-2015) is greater than 29,964 ($e^{3.4}$ = 29.964 AADT in 1000s) and the segment length is greater than 0.69 miles, the estimated number of crashes on average is 58 (Figure 4). The optimal TBR model may assign only one of the nine values (5.4, 5.1, 8.1, 22, 29, 16, 42, 20, and 58) of crashes to roadway segments based on the attributes (mean AADT, segment length, the density of major commercial driveways, and density of minor commercial driveways) selected by the optimal TBR. Note that logarithmic forms of AADT (1000s) and segment length (miles) along with other key covariates (e.g., the density of major/minor commercial driveways) in their original forms were used to train the TBR. Once the results from TBR were obtained, we took the anti-log of the values of segment length and AADT (1000s) to interpret the results – as shown in Figure 4).

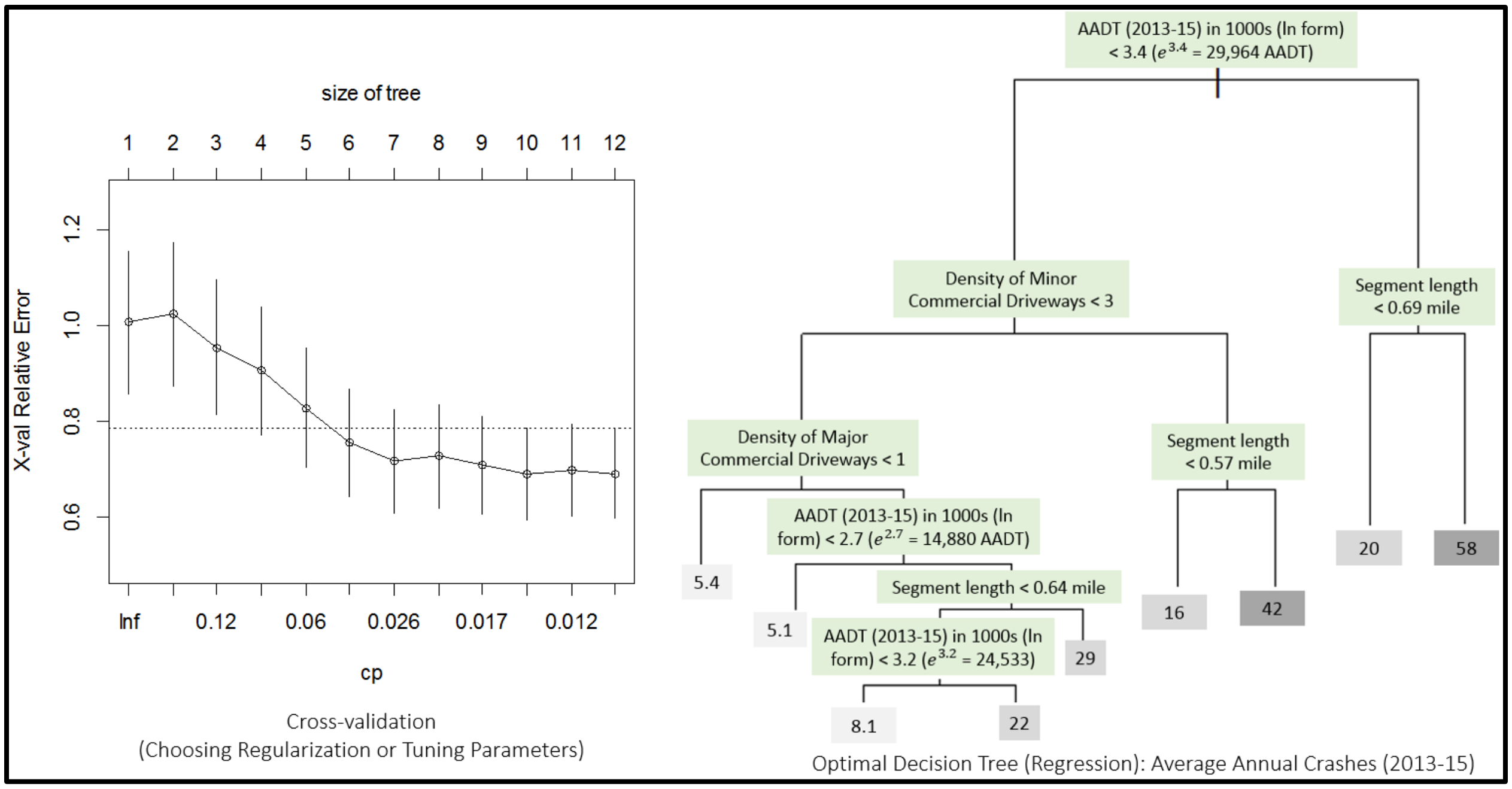

**Figure 4. Illustration of Cross validation (Regularization) and Optimal Decision-Tree Regression**

#### *3.2.2.2. Random Forest Regression*

Based on RMSE, our comprehensive grid search indicates that optimal values for the number of predictors considered in each split, number of trees, and number of maximum nodes for the final RFR are found to be 5, 250, and 14 respectively (Figure 5). Using these tuning parameters, we apply the RFR model to predict crash frequency per year using training data (Figure 5).

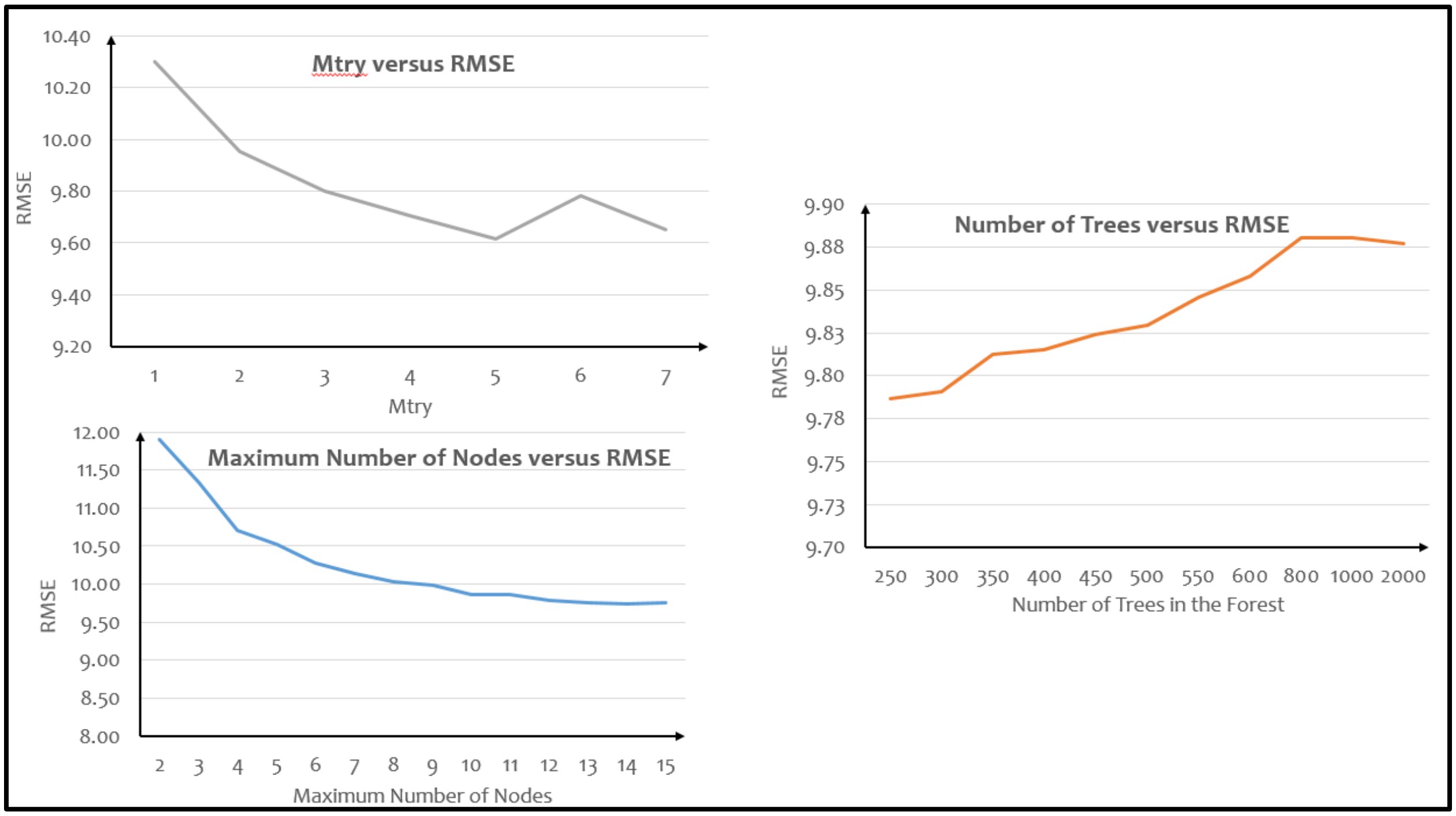

**Figure 5. Selecting Optimal Values of Regularization Parameters for Random Forest**

The relative importance of predictor variables used in the final RFR is illustrated in Figure 6. On basis of relative importance, AADT per year (2013-2015) and segment length (mile) is found to be the most important predictor variables (Figure 6). Similarly, density (number per mile) of minor commercial driveways and density of major industrial/institutional driveways are ranked at 3$^{rd}$ and 4$^{th}$ as per the final RFR model using their relative importance (Figure 6).

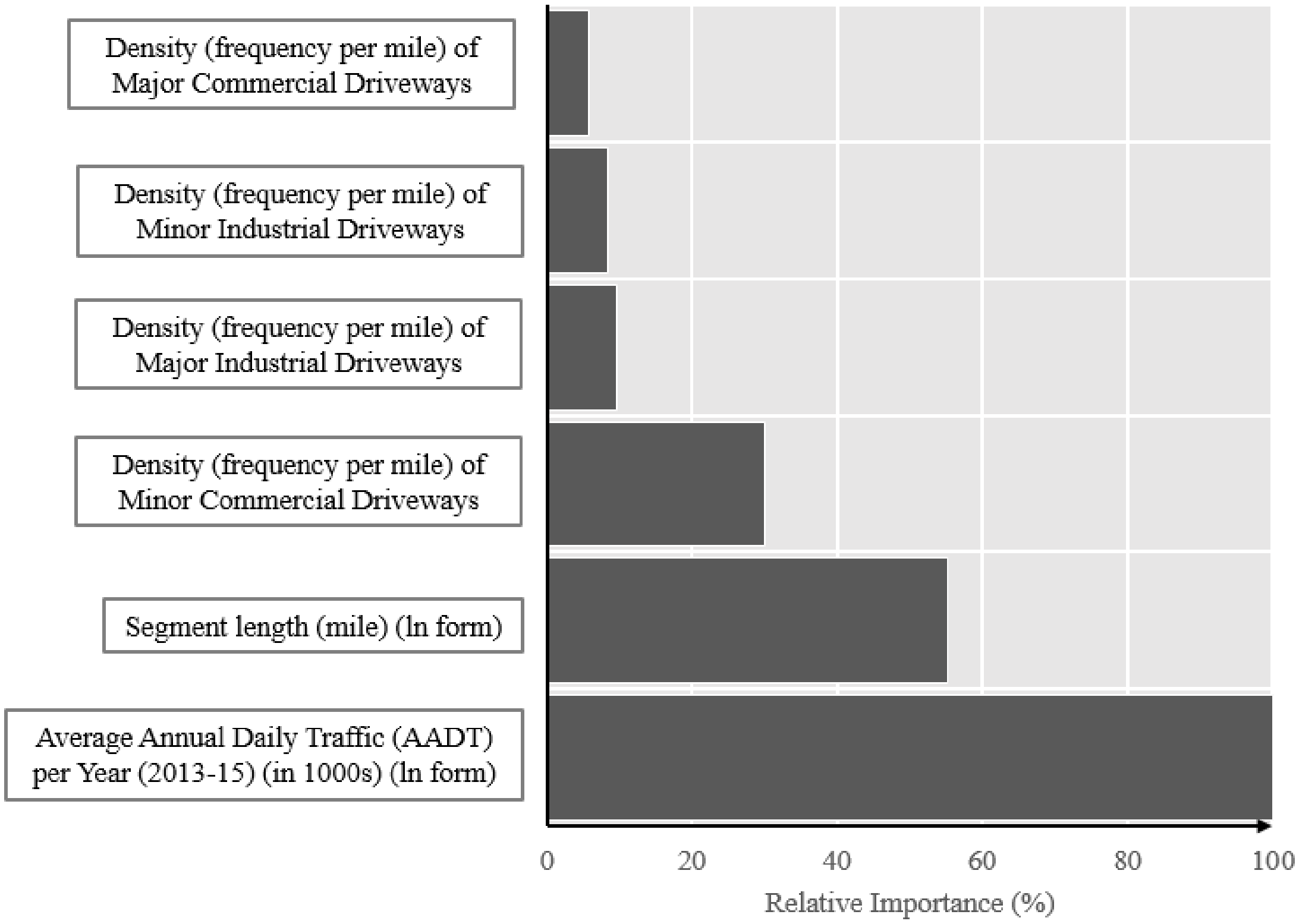

**Figure 6. Variables Relative Importance Plot: Optimal RFR (Base Learner)**

#### *3.2.2.3. Gradient Boosting Regression*

As discussed earlier, GBR is prone to overfitting, which can be minimized while achieving a balance between variance and bias through the selection of optimal regularization parameters. After conducting a grid search with all possible combinations of the number of trees, shrinkage, and interaction depth, a minimum RMSE is achieved when the number of trees, shrinkage, and complexity parameters are equal to 100, 0.1, and 3 respectively (Table 3). The performance of some key combinations of regularization parameters is shown in Table 3.

**Table 3. Selecting Optimal Combination of Regularization Parameters for Gradient Boosting**

| Shrinkage | Interaction Depth | Number of trees | RMSE | $R^2$ |
|---|---|---|---|---|
| 0.1 | 3 | 100 | 9.8098 | 0.5079 |
| 0.1 | 3 | 100 | 9.9705 | 0.5014 |
| 0.1 | 10 | 100 | 10.0821 | 0.4860 |
| 0.1 | 7 | 100 | 10.0887 | 0.4890 |
| 0.1 | 3 | 100 | 10.1875 | 0.4820 |
| 0.1 | 1 | 100 | 10.2423 | 0.4733 |

**Note:** The above six combinations are the combinations with smaller RMSE compared to all other combinations. Note that in our grid search, we assigned a range of values to shrinkage (0.1 to 1), interaction depth (1, 3, 7, and 10), and the number of trees (100, 300, 500, 1000).

Once the optimal values of the regularization are determined, a final GBR model is trained. Similar to RFR, the relative importance of key variables in predicting crash frequency per year on 5T segments of urban and suburban arterials are shown in Figure 7.

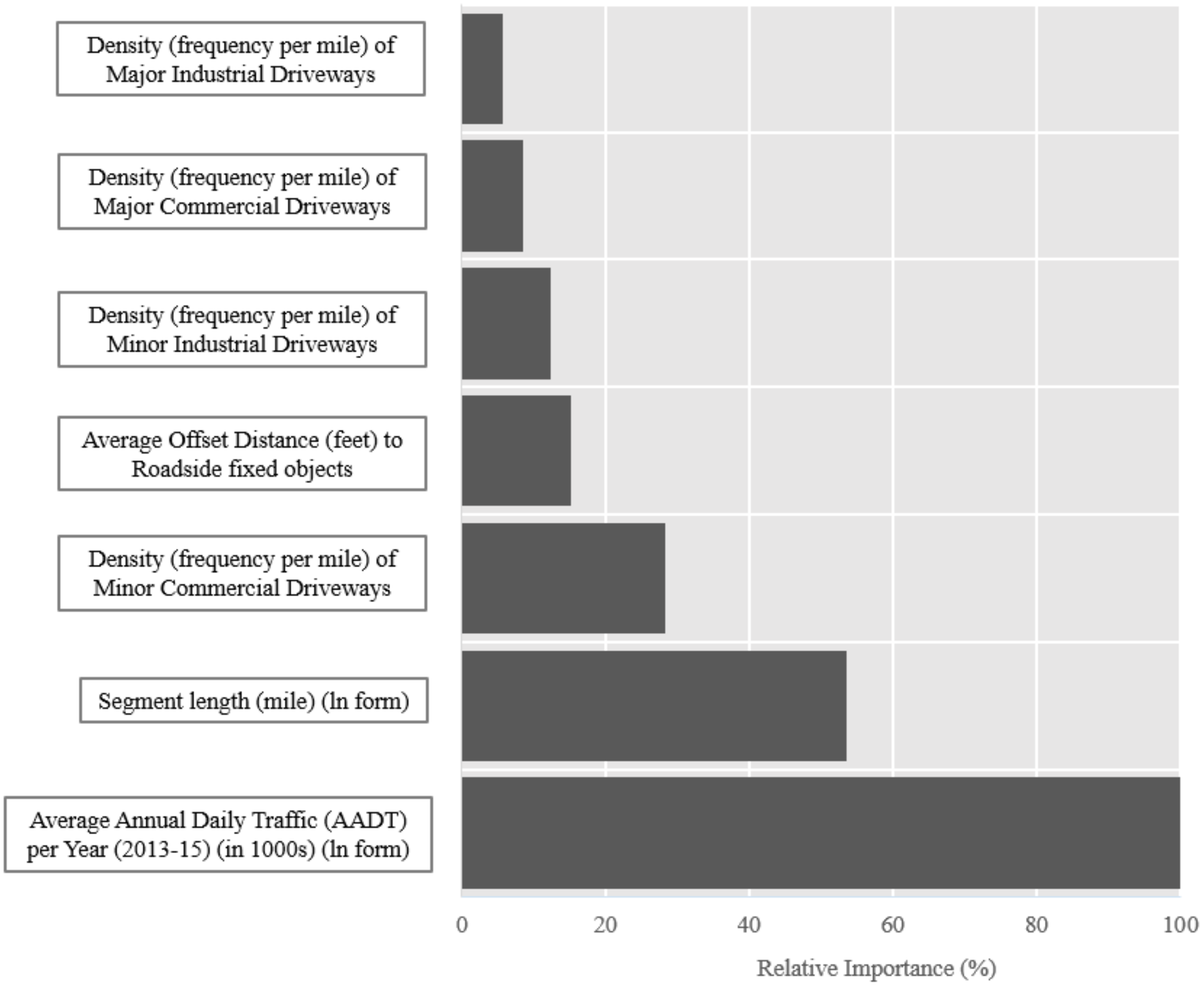

**Figure 7. Variables Relative Importance Plot: Optimal GBR (Base Learner)**

Note that GBR (the best performing base-learner) provides variable importance but does not show the magnitude or nature of the relationship between the response outcome and specific explanatory variables (*23*). We present the partial dependence plots, which are similar to MEs in statistical models, for the two key variables, AADT and segment length (Figure 8). For the sake of consistency with the best statistical model (details can be found in Section 3.2.1 and Table 2), we used the natural log forms of segment length and AADT per year (1000s), respectively in all statistical and ML base-learners. The partial dependence plots reveal a non-linear association of average yearly AADT and segment length with average yearly crash frequency (Figure 8). For instance, there is a sharp increase in crash frequency beyond an AADT of 3000 (Figure 8). With higher average yearly AADT, the frequency of average crashes (including both injury and non-injury crashes) per year increases. While previous studies reveal that total crash frequency increases with AADT (*21*), the interesting aspect of the current study is that it captures non-linearities in such a relationship through ML methods. Interestingly, if yearly AADT decreases or increases beyond the values of 9,974 and 33,115 respectively, the number of predicted crashes by optimal GBR base-learners remain constant (5 and 30 crashes per year respectively) (Figure 8). Similar interpretation applies to segment length.

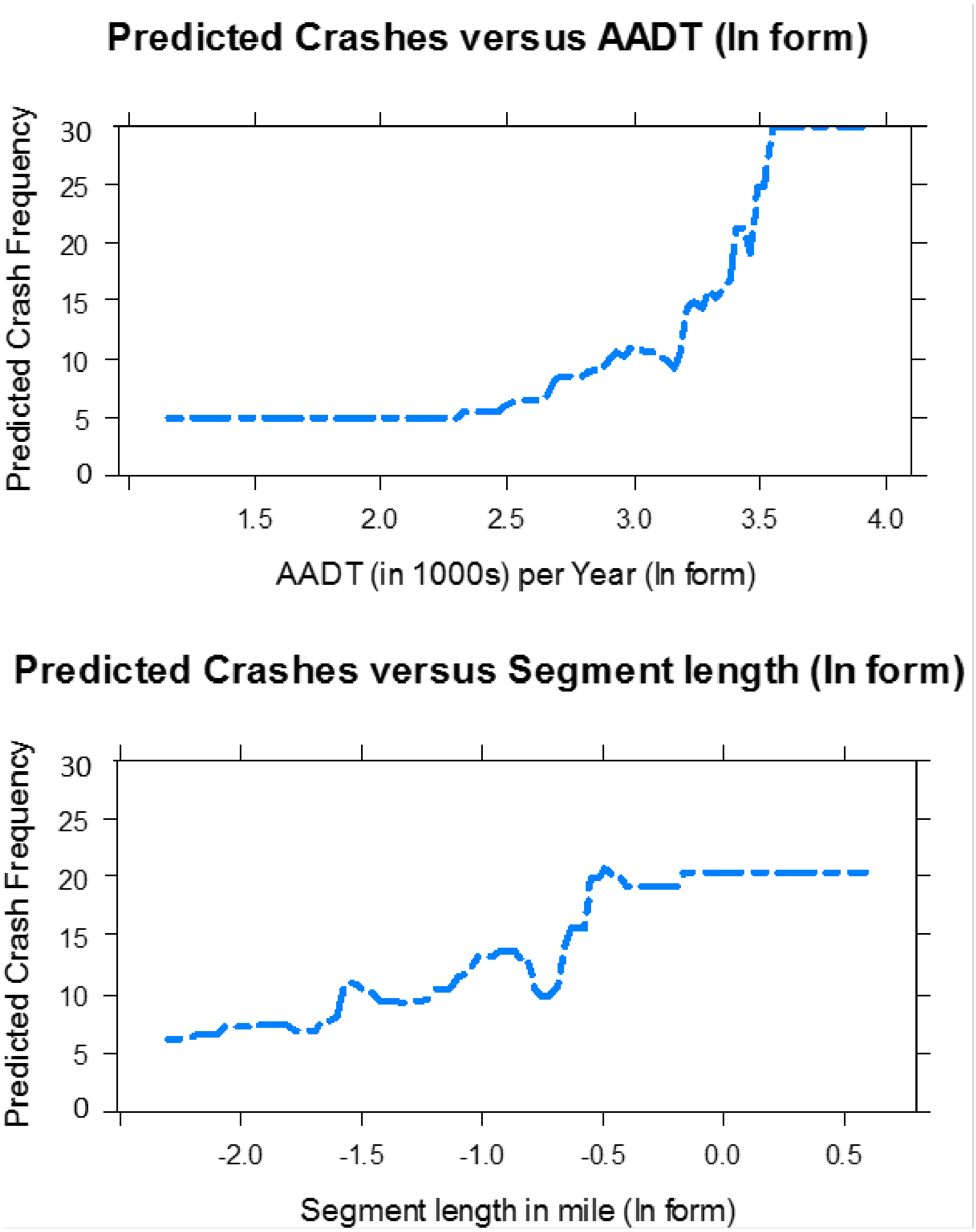

**Figure 8.** Yearly Predicted Crashes by GBR (best performing base-learner) for AADT and Segment length.

### 3.2.3. Stacking

After the five individual count data and ML-based models are developed using training data (2013-2015), the performance is evaluated using a validation dataset (2016). Next, the stacked model is trained on the validation dataset (2016) for which observed crash frequency (2016) is used as a response variable. Eventually, the predictions obtained from the five base-learners applied to the validation dataset are used as inputs (predictors) to train the stacked model. Descriptive statistics of predicted and observed crashes for the validation dataset are shown (Table 4). Note that the mean number of crashes (2016) predicted by individual models such as count data models (Poisson and Negative Binomial model) and ML models such as TBR, RFR, and GBR ($P_3$, $P_4$, and $P_5$ respectively) are very similar to the mean number of observed crashes occurred during 2016. While using the validation dataset including five new predicted values ($P_1$, $P_2$, …, $P_5$) and observed crashes, we train an RFR model as a meta-learner (stacked ensemble model) in second-stage regression. Several techniques ranging from simple linear regression to more robust ensemble methods like RFR and GBR can be used to train the stacked model.

We used the three ML methods (TBR, RFR, and GBR) as stacking meta-learners in the second-stage regression to predict crashes using the optimal combination of the base-learners. We present and discuss the results of RFR as a meta-learner (stacked ensemble method) because it led to maximum improvement in out-of-sample prediction accuracy.

Similar to individual ML models (TBR, RFR, and GBR), grid search optimization and 10-fold cross-validation procedures were used to select optimal values for regularization parameters of the Stacked model when the three methods (TBR, RFR, and GBR) were used as stacking meta-learners. The tuning parameters in the RFR used as stacked meta-learner include the number of predictors considered at each split, number of trees, and maximum number of nodes, which were found to be 2, 9, and 1000 respectively (Figure 9). To select the best number of nodes, an initial grid search was specified with a range of 2 to 15; the RMSE value was at a minimum when the maximum number of nodes equaled 9. Similarly, to select the optimal number of trees, a grid-search with the range (250, 300, 350, 400, 450, 500, 550, 600, 800, 1000, 2000) was applied which showed that 1000 trees led to the lowest RMSE (Figure 9).

**Table 4. Descriptive Statistics of Predicted and Observed Crashes (Validation Dataset: 2016)**

| Variables | Obs. | Mean | Std. Dev. | Min | Max |
|---|---|---|---|---|---|
| Total Crashes (2016) | 304 | 11.010 | 14.651 | 0.000 | 90.000 |
| Predicted Crashes via Poisson Model ($P_1$) | 304 | 11.357 | 10.869 | 0.754 | 70.664 |
| Predicted Crashes per Negative Binomial Model ($P_2$) | 304 | 11.430 | 12.119 | 0.839 | 97.733 |
| Predicted Crashes per Decision Tree Model ($P_3$) | 304 | 10.898 | 10.521 | 5.067 | 58.143 |
| Predicted Crashes per Random Forest Model ($P_4$) | 304 | 11.202 | 9.310 | 2.002 | 49.383 |
| Predicted Crashes per Gradient Boosting Model ($P_5$) | 304 | 11.345 | 11.359 | 0.000 | 53.928 |

The relative importance plot of the predictors (obtained from the five base-learners) for training the meta-learner (stacked ensemble model) is shown in Figure 10. The predicted crashes obtained from the individual RFR model ($P_4$) is found to be the most important predictor variable followed by those predicted via GBR ($P_5$) (importance = 75.45%), negative binomial model ($P_2$), and Poisson model ($P_1$) (Figure 10). Note that no weight is assigned to the predicted values by the TBR model ($P_3$) in the stacked model because there is no significant variation in $P_3$. The optimal TBR model assigns one of the 9 different values (5.4, 5.1, 8.1, 22, 29, 16, 42, 20, and 58) of crashes to segment(s) based on the attributes (shown in Figure 4) which were considered by the algorithm developing an optimal tree regression model.

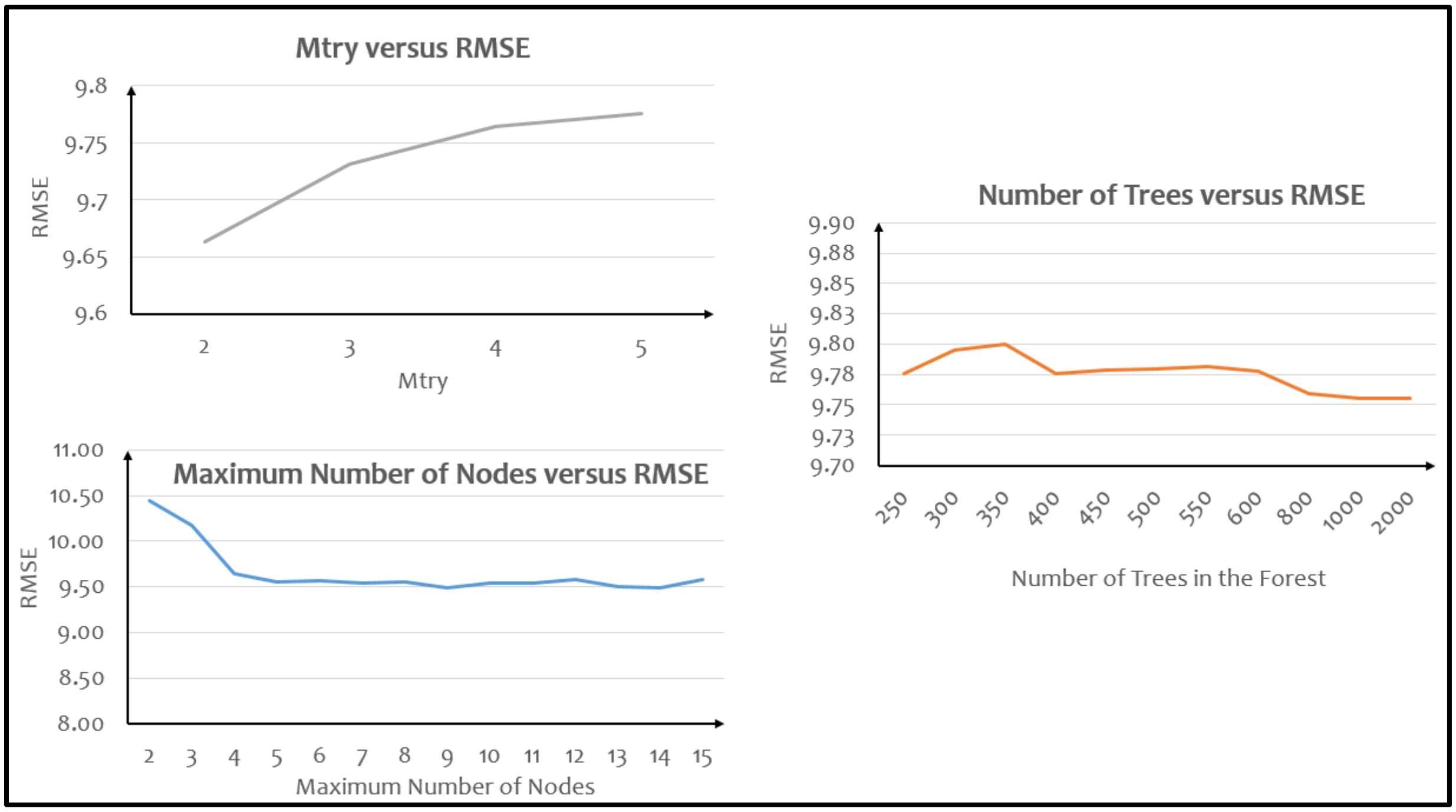

**Figure 9. Selecting Optimal Tuning Parameters for Stacked RFR Model (Second-Stage Regression)**

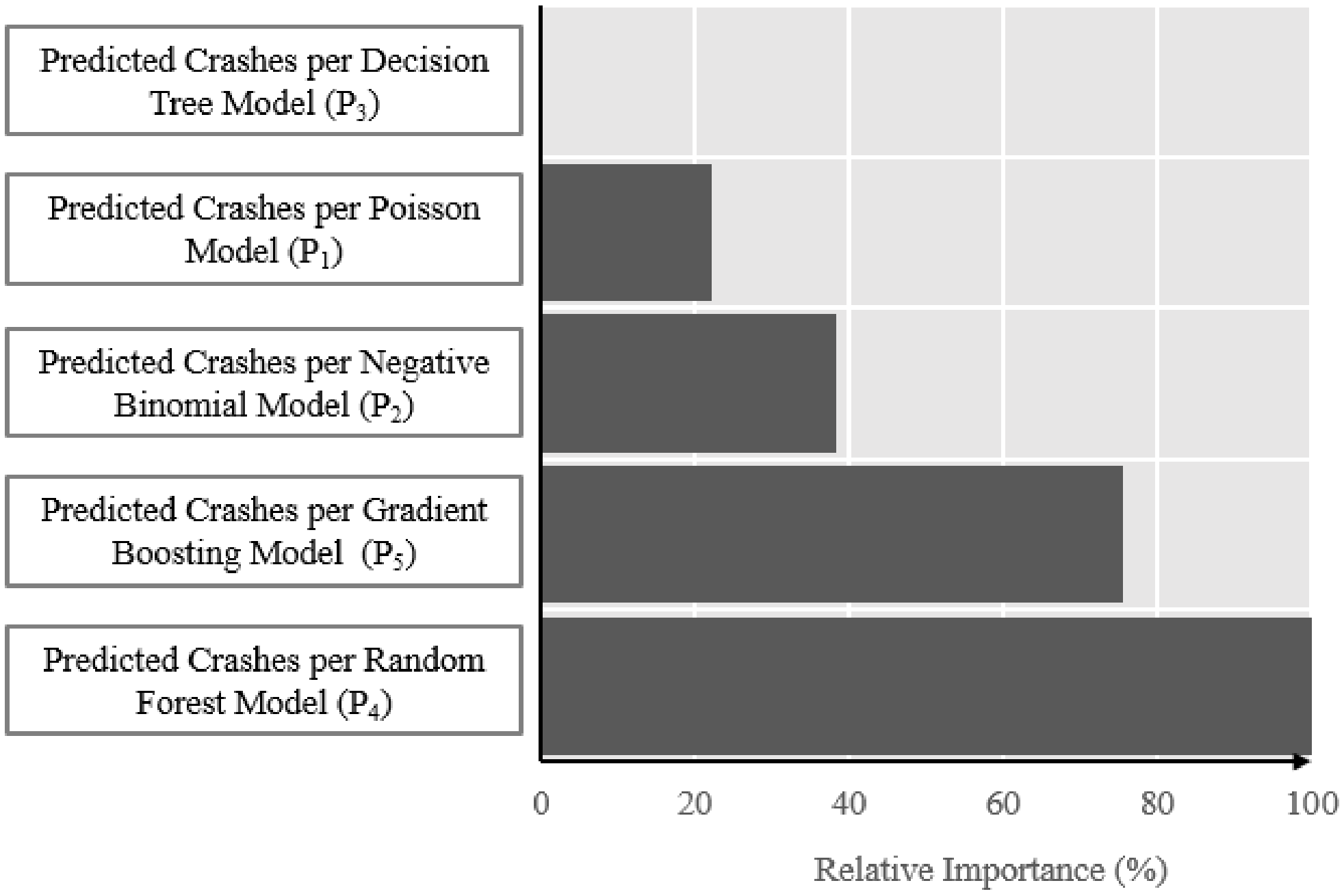

**Figure 10. Variables Relative Importance Plot: Optimal Stacked RFR Model (Meta Learner)**

### 3.2.4. Comparing Out-of-Sample Prediction Performance

To compare the predictive performance of the five base-learners and meta-learners based on the new dataset (neither used to train base-learners nor the meta-learner), we computed out-of-sample RMSE and MAE (Table 5). Our findings indicate that GBR has the lowest out-of-sample RMSE and MAE among all base-learners (Table 5). Referring to the predictive performance of meta-learners, both RFR and GBR as stacking meta-learners further reduced out-of-sample RMSE and MAE compared to the best performing base-learner (GBR) (Table 5). To conclude, RFR, as a stacking meta-learner, is found to have the lowest out-of-sample RMSE and out-of-sample MAE among all the base-learners and meta-learners and is selected as the best performing model for out-of-sample crash prediction (Table 5). For the sake of brevity, we only discuss the results of RFR as a stacking meta-learner in the paper.

To have a deeper understanding of the out-of-sample prediction errors, we also provide distributional statistics of out-of-sample absolute prediction error (Table 5). RFR as a stacking meta-learner leads to the lowest out-of-sample absolute prediction error (Table 5). The standard deviations of absolute prediction errors for GBR and RFR as stacking meta-learners are found to be the lowest indicating less spreading-out around the mean value of the error (Table 5).

**Table 5. Comparison Prediction Performance (Out-of-Sample): RMSE and MAE**

| **Comparing Out-of-Sample Prediction Accuracy (RMSE & MAE)** | | | | | | **Distribution of Absolute Prediction Error**<br>**Absolute (observed crashes - predicted crashes)** | | | | |
|---|---|---|---|---|---|---|---|---|---|---|
| **Model** | Type | RMSE | % Difference in RMSE compared to GBR Model* | MAE | % Difference in MAE compared to GBR Model** | **N** | **Mean** | **Std. Dev.** | **Minimum** | **Maximum** |
| TBR | Meta Learner | 9.515 | 6.98 | 5.889 | 0.43 | 304 | 5.889 | 7.487 | 0.119 | 81.000 |
| Gradient Boosting | Meta Learner | 8.404 | -5.51 | 5.609 | -4.33 | 304 | 5.609 | 6.268 | 0.015 | 59.010 |
| Random Forest | Meta Learner | 8.312 | -6.54 | 5.383 | -8.19 | 304 | 5.383 | 6.345 | 0.043 | 62.137 |
| Poisson | Base Learner | 10.123 | 13.82 | 6.315 | 7.71 | 304 | 6.315 | 7.926 | 0.033 | 77.058 |
| Negative Binomial | Base Learner | 11.118 | 25.01 | 6.589 | 12.38 | 304 | 6.589 | 8.970 | 0.008 | 82.822 |
| TBR | Base Learner | 10.251 | 15.26 | 6.757 | 15.25 | 304 | 6.757 | 7.722 | 0.038 | 78.333 |
| Random Forest | Base Learner | 9.023 | 1.45 | 5.951 | 1.50 | 304 | 5.951 | 6.794 | 0.046 | 71.128 |
| Gradient Boosting | Base Learner | 8.894 | Base | 5.863 | Base | 304 | 5.863 | 6.698 | 0.000 | 68.084 |

**Note: *** % Difference in RMSE compared to GBR Model $= \frac{(\text{RMSE}_{Model\ X} - \text{RMSE}_{GBR\ Model})}{\text{RMSE}_{GBR\ Model}} * 100\%$

** % Difference in MAE compared to GBR Model $= \frac{(\text{MAE}_{Model\ X} - \text{MAE}_{GBR\ Model})}{\text{MAE}_{GBR\ Model}} * 100$

To visualize and compare the out-of-sample prediction performance of base-learners and the stacked ensemble technique (meta-learner), we present plots of predicted versus observed crashes based on the testing data (2017) (Figure 11). The RFR ensemble model, when used as a meta-learner, shows the best fit, followed by when GBR is used as a stacking meta-learner, GBR (base learner), and RFR (base learner) as shown in Figure 11. Similar findings were obtained in other fields where prediction accuracy for the stacked ensemble model (used for classification) improved by 2%-4% (*15*). To conclude, we found that the application of the stacked ensemble technique can help in obtaining more accurate crash predictions in the future. Note that none of the studies have evaluated the applicability of more accurate, reliable, and intelligent heterogeneous ensemble procedures to determine the crash frequency.

The plot of predicted versus actual crash frequency for the TBR models (when used as a base- or stacking meta learner) seem unusual compared to the other models. In our case, the optimal TBR (when used as a base learner) model indicates that one of the nine values (5.4, 5.1, 8.1, 22, 29, 16, 42, 20, and 58) of crashes may be assigned to any segment based on its attributes (mean AADT, segment length, the density of major commercial driveways, and density of minor commercial driveways). Similar predictions can be seen in Figure 11 when TBR is used as a stacked model.

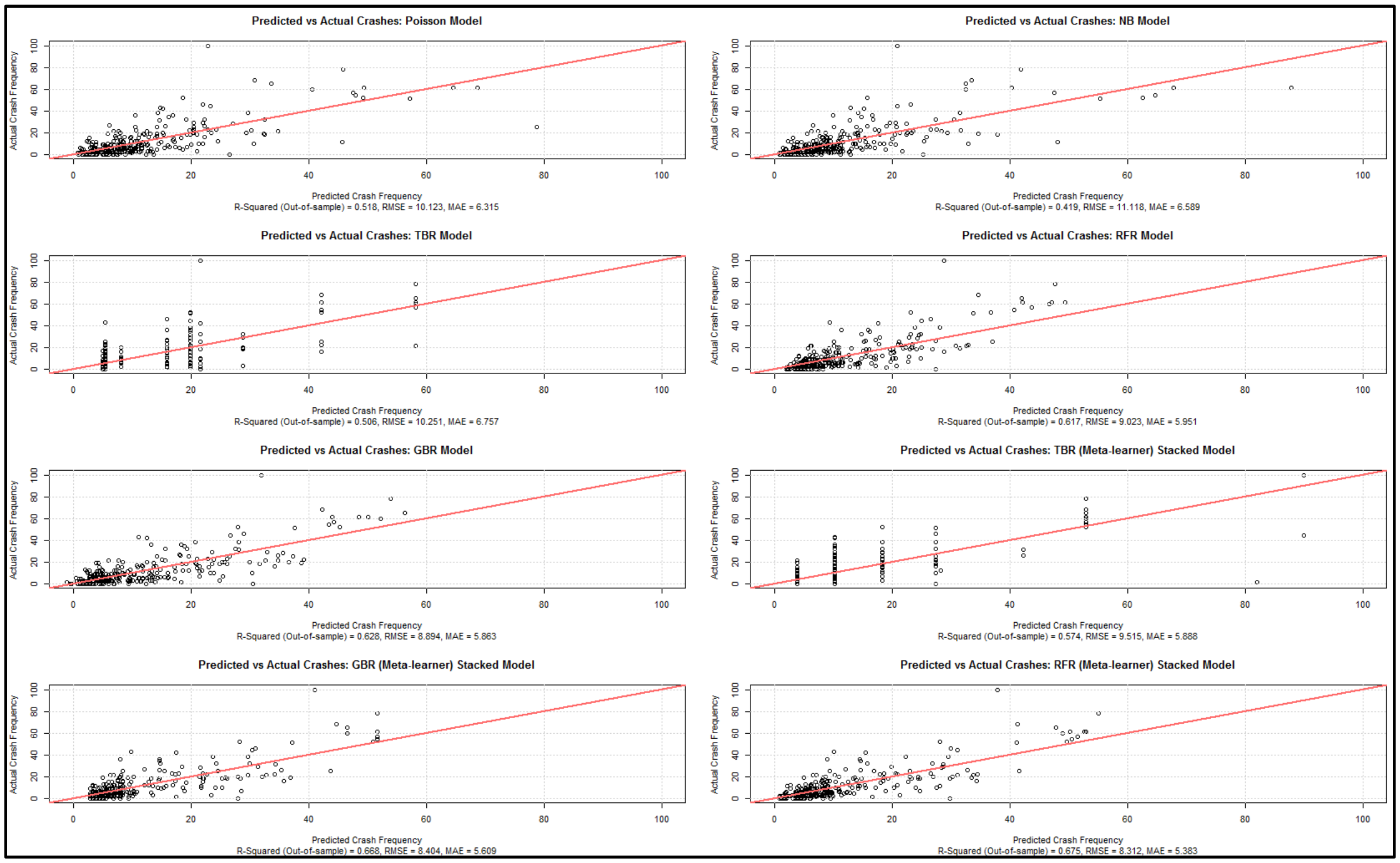

**Figure 11. Out-of-sample prediction: Observed versus Predicted Crashes**

## 4. Limitations and Future Directions

This study uses 5T segments of urban and suburban arterials in Tennessee and may not be extended to other states due to variations in driving behavior, socio-demographic, and roadway conditions. While stacking may significantly improve the out-of-sample prediction accuracy, it does not provide the variable importance for the actual predictor variables (e.g., segment length and AADT). Note that in this study, stacking is applied to combine multiple predictions (as opposed to combining distributions of coefficients, such as in the Bayesian setup). Thus, the inference is not relevant in Stage-2. However, the inferences are provided in Stage-1 by individual base-learners that include statistical and ML models. Based on our study objectives, we split five years of crash data into training (2013-2015), validation (2016), and testing (2017) datasets - indicting that only crash frequency and AADT may vary across the datasets while roadway geometry remains similar. In the future, data splitting can also be done using standard splitting procedures rather than the year-wise split, depending on study design and objectives. The application presented herein is based on year-wise splits, assuming temporal transferability of the models over years. As part of future work, variants of the methods presented herein that relax this assumption can be examined.

## 5. Conclusions

Traditional count data models and ML methods have been extensively used in the safety literature for the development of statistical relationships between crash frequency and associated factors. This study contributes by presenting a rigorous and novel heterogeneous ensemble methods (HEM) scheme to “stack” predictions from competing frequentist and ML models – eventually leading to a more accurate prediction of crashes. By using a more accurate and reliable intelligent pattern recognition scheme, the “Stacking” methodology harnesses the inferential framework provided by traditional count data models and the predictive power offered by ML methods. The objectives are achieved using five-year crash, traffic, and roadway geometric data for 5T urban and suburban arterials in Tennessee. To the best of the authors’ knowledge, no study to date has applied heterogeneous ensemble methods to pool multiple predictions from frequentist and ML methods.

The results suggest the significant potential of “Stacking” in providing more accurate predictions by heterogeneously assembling crash forecasts from individual statistical (Poisson and negative binomial) and machine-learning-based base-learners (TBR, RFR, and GBR). Using out-of-sample prediction performance, the GBR model led to the lowest RMSE and MAE values among all the individual base-learners. While individual ML-based base learners can provide greater predictive accuracy, there is no escaping the relationship between bias and variance underpinning most ML models. In other words, using a single supervised or unsupervised ML method could lead to relatively less accurate predictions due to the compromised bias or variance. By superimposing an ML-based meta learner on predictions obtained from the five statistical and ML-based base-learners, the RMSE and MAE values of crash forecasts were further reduced by 6.54% and 8.19% respectively compared to the prediction accuracy of the best-fit GBR based individual base-learner. From an inferential standpoint, the individual base-learners offer insights into the links between crash frequency and associated factors. Count data models show that besides exposure variables (AADT and segment length), higher accessibility correlates with higher crash frequency. Contrarily, a larger offset distance to a fixed object correlates with lower crash frequency. In terms of variable importance, the three ML-based base-learners rank AADT, segment length, and density of minor commercial driveways as the three top predictors of crash frequency.

The results of this study have important implications. By using heterogeneous ensemble methods such as Stacking, even more, accurate crash forecasts can be obtained compared to those obtained from individual frequentist or ML methods. With more accurate crash forecasts, roadway segments can be better prioritized in terms of the need for place-based safety countermeasures. From a practical standpoint, the straightforward heterogeneous ensemble method technique can be easily automated for more accurate crash prediction. From a research perspective, the methodology can be expanded by other researchers to include an even broader set of ML methods or considering more rigorous simulation-assisted statistical methods accounting for methodological issues like observed and unobserved heterogeneity.

## 6. Acknowledgments

This study was funded by the Tennessee Department of Transportation (Grant No. RES2020-04) and the U.S. DOT through the Collaborative Sciences Center for Road Safety; a consortium led by The University of North Carolina at Chapel Hill in partnership with The University of Tennessee Knoxville. The views in this paper are solely those of the authors who are responsible for the content of this publication. The authors also wish to sincerely thank Amin Mohammadnazar (Graduate Student in Civil Engineering at University of Tennessee, Knoxville) for his help in collecting a portion of the data used in this study. The authors also extend their sincere gratitude to Ms. Meredith King, an undergraduate student in Civil and Environmental Engineering at the University of Tennessee (Knoxville), for proofreading the article.

## Author Contributions

The authors confirm contribution to the paper as follows: study conception and design: Numan Ahmad, Behram Wali, Asad Khattak; data collection: Numan Ahmad; analysis and interpretation of results: Numan Ahmad, Behram Wali; draft manuscript preparation Numan Ahmad, Behram Wali, Asad Khattak. All authors reviewed the results and approved the final version of the manuscript.

**References**

[1] Shankar, V., F. Mannering, and W. Barfield. Effect of roadway geometrics and environmental factors on rural freeway accident frequencies. *Accident Analysis & Prevention,* Vol. 27, No. 3, 1995, pp. 371-389.

[2] Khattak, A., N. Ahmad, A. Mohammadnazar, I. MahdiNia, B. Wali, and R. Arvin. Highway Safety Manual Safety Performance Functions & Roadway Calibration Factors: Roadway Segments Phase 2, Part. 2020.

[3] Srinivasan, R., and D. Carter. Development of safety performance functions for North Carolina.In, North Carolina. Dept. of Transportation. Research and Analysis Group, 2011.

[4] Pan, G., L. Fu, and L. Thakali. Development of a global road safety performance function using deep neural networks. *International Journal of Transportation Science and Technology,* Vol. 6, No. 3, 2017, pp. 159-173.

[5] Saha, D., P. Alluri, and A. Gan. Prioritizing Highway Safety Manual's crash prediction variables using boosted regression trees. *Accident Analysis & Prevention,* Vol. 79, 2015, pp. 133-144.

[6] Breiman, L., J. Friedman, C. J. Stone, and R. A. Olshen. *Classification and regression trees*. CRC press, 1984.

[7] Thapa, R., S. Gupta, A. Gupta, D. Reddy, and H. Kaur. Use of geospatial technology for delineating groundwater potential zones with an emphasis on water-table analysis in Dwarka River Basin, Birbhum, India. *Hydrogeology Journal,* Vol. 26, No. 3, 2018, pp. 899-922.

[8] Bhatt, S., E. Cameron, S. R. Flaxman, D. J. Weiss, D. L. Smith, and P. W. Gething. Improved prediction accuracy for disease risk mapping using Gaussian process stacked generalization. *Journal of The Royal Society Interface,* Vol. 14, No. 134, 2017, p. 20170520.

[9] Sabzevari, M., G. Martínez-Muñoz, and A. Suárez. Pooling homogeneous ensembles to build heterogeneous ones. *arXiv preprint arXiv:1802.07877*, 2018.

[10] Chali, Y., S. A. Hasan, and M. Mojahid. Complex question answering: homogeneous or heterogeneous, which ensemble is better? In *International Conference on Applications of Natural Language to Data Bases/Information Systems*, Springer, 2014. pp. 160-163.

[11] Fernández-Alemán, J. L., Carrillo-de-Gea, J. M., Hosni, M., Idri, A., and García-Mateos, G. Homogeneous and heterogeneous ensemble classification methods in diabetes disease: a review.In *2019 41st Annual International Conference of the IEEE Engineering in Medicine and Biology Society (EMBC)*, IEEE, 2019. pp. 3956-3959.

[12] Farid, A., M. Abdel-Aty, and J. Lee. Comparative analysis of multiple techniques for developing and transferring safety performance functions. *Accident Analysis & Prevention,* Vol. 122, 2019, pp. 85-98.

[13] Iranitalab, A., and A. Khattak. Comparison of four statistical and machine learning methods for crash severity prediction. *Accident Analysis & Prevention,* Vol. 108, 2017, pp. 27-36.

[14] Tang, J., J. Liang, C. Han, Z. Li, and H. Huang. Crash injury severity analysis using a two-layer Stacking framework. *Accident Analysis & Prevention,* Vol. 122, 2019, pp. 226-238.

[15] Güneş, F., R. Wolfinger, and P.-Y. Tan. Stacked ensemble models for improved prediction accuracy.In *Proc. Static Anal. Symp.*, 2017. pp. 1-19.

[16] Van der Laan, M. J., E. C. Polley, and A. E. Hubbard. Super learner. *Statistical Aapplications in Genetics and Molecular Biology,* Vol. 6, No. 1, 2007.

[17] Anastasopoulos, P. C., and F. L. Mannering. A note on modeling vehicle accident frequencies with random-parameters count models. *Accident Analysis & Prevention*, Vol. 41, No. 1, 2009, pp. 153-159.

[18] Washington, S., Karlaftis,. MG, and Mannering, FL. Statistical and Econometric Methods for Transportation Data Analysis. In, Chapman & Hall/CRC New York, 2010.

[19] Breiman, L. Bagging predictors. *Machine Learning,* Vol. 24, No. 2, 1996, pp. 123-140.

[20] Hastie, T., R. Tibshirani, and J. Friedman. *The elements of statistical learning: data mining, inference, and prediction*. Springer Science & Business Media, 2009.

[21] AASHTO. Highway Safety Manual. *Washington, DC, No. 529*, **2010**.

[22] Albuquerque, F. D., and D. M. Awadalla. Roadside design assessment in an urban, low-density environment in the Gulf Cooperation Council region. *Traffic Injury Prevention,* Vol. 20, No. 4, 2019, pp. 436-441.

[23] Friedman, J. H. Greedy function approximation: a gradient boosting machine. *Annals of Statistics*, 2001, pp. 1189-1232.